# Real-time monitoring of driver drowsiness on mobile platforms using 3D neural networks

Jasper S. Wijnands[1] · Jason Thompson[1] · Kerry A. Nice[1] · Gideon D. P. A. Aschwanden[1] · Mark Stevenson[1,2,3]

**Abstract**

Driver drowsiness increases crash risk, leading to substantial road trauma each year. Drowsiness detection methods have received considerable attention, but few studies have investigated the implementation of a detection approach on a mobile phone. Phone applications reduce the need for specialised hardware and hence, enable a cost-effective roll-out of the technology across the driving population. While it has been shown that three-dimensional (3D) operations are more suitable for spatiotemporal feature learning, current methods for drowsiness detection commonly use frame-based, multi-step approaches. However, computationally expensive techniques that achieve superior results on action recognition benchmarks (e.g. 3D convolutions, optical flow extraction) create bottlenecks for real-time, safety-critical applications on mobile devices. Here, we show how depthwise separable 3D convolutions, combined with an early fusion of spatial and temporal information, can achieve a balance between high prediction accuracy and real-time inference requirements. In particular, increased accuracy is achieved when assessment requires motion information, for example, when sunglasses conceal the eyes. Further, a custom TensorFlow-based smartphone application shows the true impact of various approaches on inference times and demonstrates the effectiveness of real-time monitoring based on out-of-sample data to alert a drowsy driver. Our model is pre-trained on ImageNet and Kinetics and fine-tuned on a publicly available Driver Drowsiness Detection dataset. Fine-tuning on large naturalistic driving datasets could further improve accuracy to obtain robust in-vehicle performance. Overall, our research is a step towards practical deep learning applications, potentially preventing micro-sleeps and reducing road trauma.

✉ Jasper S. Wijnands
jasper.wijnands@unimelb.edu.au

1 Transport, Health and Urban Design Research Hub, Melbourne School of Design, The University of Melbourne, Parkville, VIC 3010, Australia

2 Melbourne School of Engineering, The University of Melbourne, Parkville, VIC 3010, Australia

3 Melbourne School of Population and Global Health, The University of Melbourne, Parkville, VIC 3010, Australia

## 1 Introduction

Each year, motor vehicle accidents contribute to over 1.2 million fatalities globally [52]. In 95–99% of these crashes, human error, including driver drowsiness, is a contributing factor [14]. For example, in the USA, crashes related to driver fatigue led to over 800 fatalities in 2014 and 37,000 injuries per year between 2005 and 2009 [29].

The association between driver drowsiness and crash risk has been confirmed in various studies. For example, Williamson et al. [55] found that sleep homeostatic effects produce impaired performance and accidents. Klauer et al. [24] identified that drowsy drivers are four to six times more likely to be involved in a crash or near-crash than attentive drivers. Cumulative sleep debt also increases the risk of crashing, as shown in a case-control study of heavy-vehicle drivers [43]. Bouchner et al. [5] showed that drowsy drivers demonstrated more fast corrective movements of the steering wheel, larger deviations from the ideal trajectory and were less likely to adhere to the speed limit.

Various approaches have been implemented to reduce road trauma, such as educating drivers in improved fatigue management practices; e.g. taking sufficient rest breaks [11]. This relies on self-assessment of a driver's drowsiness levels, and previous research has shown that drivers can indeed identify their current state of sleepiness and likelihood of falling asleep [56]. Although self-assessment of drowsiness provides a good initial coping mechanism, it does not fully eliminate drowsiness-related road trauma, so there is a need to develop complementary warning and safety systems.

## 1.1 Drowsiness detection

Technological innovations have the potential to further reduce the number of injuries and fatalities among road users by alerting drivers when a drowsy state has been detected. Nåbo [28], for example, found that drowsiness warnings resulted in drivers looking at the road more quickly than when they had to recognise concentration loss themselves. Further, Blommer et al. [4] showed that driver reaction times with respect to lane departure improved when a warning was given. The manner in which these warnings were provided was not crucial: visual, haptic and sound warnings were all equally effective.

Published drowsiness detection methodologies can be split into various categories, including vehicle-based measurements, physiological measurements, and computer vision techniques.

### 1.1.1 Vehicle-based measurements

These approaches generally focus on monitoring steering wheel movements (e.g. [38]) and deviations from lane position (e.g. [26]) to detect large or sudden corrections in travelling direction.

### 1.1.2 Physiological measurements

An alternative method to detect changes in alertness is by monitoring internal signals, such as brain activity (e.g. [23]) or heart rate variability (e.g. [46]). However, because the driver is required to wear various sensors, physiological measurements are often impractical when compared to vehicle-based or computer vision solutions.

### 1.1.3 Computer vision techniques

Artificial neural networks are increasingly being applied to road safety issues (e.g. toll plaza safety [2], finding patterns in driving behaviour [53]) and are currently also the leading method for image recognition tasks [39]. This approach for drowsiness detection uses video of drivers' faces and has been used by various studies; an overview is provided in Table 1.

Generally, the published methods based on computer vision use a multi-step approach, with a separate step to identify features such as eyelid closure and head position. Vural et al. [49] identified various other facial expressions important for drowsiness prediction by data mining human behaviour. The most predictive expressions for drowsiness include outer brow raise, frowning, chin raise and nose wrinkle. Some of these expressions may be indicative of early stages of fatigue (e.g. outer brow raises in an attempt to keep the eyes open). However, many existing methods for drowsiness detection focus solely on (a selection of) eye, mouth and head position. Further, some methods use neural network architectures which were developed for image recognition. Although previous methods have proved successful, there are limitations associated with applying architectures designed for two-dimensional (2D) applications to detection issues that contain a third dimension of time (e.g. drowsiness detection). For example, closed eyes in a single frame could indicate both blinking (i.e. not drowsy) or a micro-sleep (i.e. drowsy). Similarly, it may be difficult to distinguish between talking and yawning. The analysis of single frames also cannot identify face movements that are important for drowsiness detection (e.g. nodding vs. looking in side mirrors). Therefore, an additional step is generally implemented to link identified features to drowsiness. The integration of spatial and temporal information in a single model presents opportunities to further improve drowsiness detection methodologies. Spatiotemporal modelling has also been applied successfully in other domains to provide (real-time) warnings; e.g. flood warnings through river flow monitoring [57, 58]. Further, the spatiotemporal modelling of human actions has been investigated extensively in the field of action recognition.

## 1.2 Action recognition on video benchmarks

Besides the specific application to drowsiness detection, various studies have explored action recognition in general. The performance of new action recognition methods is generally assessed using video benchmarks, for example, UCF-101 [42], HMDB-51 [25] and Kinetics [22]. These experiments have led to various insights on techniques that provide superior performance for action recognition. For example, three-dimensional (3D) convolutions outperform approaches based on hand-crafted features (i.e. as shown using the TRECVID benchmark) [18]. Further, Tran et al. [45] and Karpathy et al. [21] showed that 3D convolutional neural networks are more suitable for learning spatiotemporal features from video than 2D networks. Other successful approaches include the extraction of motion information before feeding it into a neural network. Simonyan and Zisserman [41] showed how optical flow extraction as a pre-processing step leads to higher performance than using video data directly. However, the highest accuracy was achieved by combining a spatial and a temporal network through fusion of the outputs. This indicates that appearance and motion contain complementary information and combining them can improve performance on action recognition tasks. Feichtenhofer et al. [10] further improved this approach by connecting the spatial and temporal networks in intermediate layers.

**Table 1** Overview of existing studies using computer vision techniques for drowsiness detection

| Study | Methodology |
| --- | --- |
| Park et al. [30] | Three pre-trained deep neural networks (AlexNet, VGG-FaceNet and FlowImageNet) along with two ensemble strategies (independently averaged architecture and feature-fused architecture) classify each video frame as drowsy or not |
| Jiménez et al. [20] | Haar classifiers detect head, eye and mouth segments in video frames; a neural network then quantifies the level of driver distraction in each frame |
| Ying et al. [59] | Colour detection methods locate the face, mouth and eyes of the driver, followed by a three-layered neural network to assess states of the eyes (closed, open, narrow) and mouth (closed, open normally, open widely). The monitoring system can provide several types of warnings |
| Huynh et al. [16] | A face tracking algorithm clips each video frame and feeds it to a 3D convolutional neural network. A boosting technique combined with semi-supervised learning using the validation set further enhances accuracy |
| Ribarić et al. [34] | Evaluates head rotations and eye and mouth openness, while a knowledge-based decision model decides whether to issue a warning or alarm based on the detected drowsiness level |
| Ji et al. [17] | Various algorithms track the driver's head and eyelid movement, gaze and facial expression; a Bayesian network assesses fatigue using these features |
| Jiangwei et al. [19] | Detects mouth movements as a single feature for a three-layered neural network, classifying each frame as dozing (mouth wide open), talking (moderately open) or silent (mouth closed) |
| Rong-ben et al. [35] | Detects drowsiness by focussing purely on the driver's eyes |
| Flores et al. [12] | Machine learning techniques detect a driver's face and eyes in a single frame; these features are tracked over time by a neural network. The frequency of eye blinks is used as a drowsiness indicator, while head position is used as an indicator for distraction |
| Lenskiy and Lee [27] | A neural network and facial segmentation algorithm obtain the driver's facial features, which are then used to track the iris and detect blinking. Eye closures longer than 220 ms are classified as drowsy |
| Harada et al. [13] | Calculates pupil diameter from eye-tracking data, which is used in a recurrent neural network to predict a driver's distraction level |

To reduce the computational requirements of these high-performing neural networks while limiting reductions in classification accuracy, several low-power architectures have been developed (e.g. [15, 37, 60, 61]). These architectures replace parts of the network with high computational requirements by new designs, such as depthwise separable and pointwise group convolutions. The networks are specifically designed for image recognition on mobile devices and require 2D inputs. As action recognition may have higher computational requirements, deploying these types of methods on mobile devices is complex.

Beyond action recognition, activity prediction [36] (also referred to as human intention inference) concerns the early recognition of unfinished activities instead of the after-the-fact classification of completed activities. Intention inference only needs a partial rather than a complete sequence of observations [9]. An example of activity prediction is the assessment of drowsiness based on short samples, where frames only contain the onset of a yawning event.

In summary, the gap in existing research is how to incorporate some of the action recognition techniques successfully deployed in high-performance systems for use on mobile platforms, in order to take advantage of associated improvements in prediction accuracy. This paper describes a new method to directly detect drowsiness in drivers based on activity prediction from real-time face video using depthwise separable 3D convolution operations. Applying this methodology to real-time video of the driver can potentially identify micro-sleeps and alert drivers, with an ultimate outcome of reducing fatigue-related road crashes and road trauma.

## 2 Methods

### 2.1 Initial considerations for mobile deployment

This section provides considerations for implementation of a drowsiness detection model on mobile or in-vehicle devices. For safety-critical applications, it is important that the real-time assessment is timely. The task requirements for such a system, which will be discussed below, include (1) the collection of frames, (2) pre-processing recorded information, (3) model inference and (4) results processing. The duration of the modelled action is therefore an important consideration. Varol et al. [47] show that incorporating more consecutive input frames leads to higher performance on UCF-101. However, capturing more frames than strictly necessary will hinder real-time detection, as both video recording time and model inference time increase. Hence, it is important to understand the duration of symptoms related to driver drowsiness to maximise detection accuracy. In this research, we capture ten frames at a frame rate of 30 frames per second (fps), leading to 333 ms of video recording. Secondly, computationally expensive pre-processing steps should be avoided where possible. For example, as described above, previous studies have shown improved performance following motion extraction through optical flow computation. Although optical flow extraction as a pre-processing step may increase accuracy, it is not necessarily optimal for mobile platforms, since real-time extraction has considerable additional computational requirements. Therefore, we do not use optical flow for mobile deployment. Thirdly, the neural network architecture has a large influence on mobile run-time. One option is model deployment on custom hardware optimised for inference; few or no changes to the network structure are required in that case. This approach is suitable for in-vehicle solutions, using a built-in camera and high-speed graphics processing unit (GPU) computing modules for model inference. For deployment on mobile phones or other mobile devices not specifically developed for artificial intelligence, computation speed for inference is the limiting factor. For example, early experiments on a Raspberry Pi 3 yielded inference times over 10 s using a fairly compact network architecture; not fast enough to provide real-time warnings to drivers experiencing a micro-sleep. The alternative option, used in this research, is therefore to use an optimised network structure designed for mobile deployment. The final step of the drowsiness detection system is results processing, which generally only takes up a small portion of total run-time.

## 2.2 Data

An academic Driver Drowsiness Detection (DDD) dataset [51] was used, first introduced during the 2016 Asian Conference on Computer Vision. As this DDD dataset is also used by various other studies (e.g. [7, 30, 32]), it allows for an easy comparison of drowsiness detection methodologies. Videos were recorded at a $480 \times 640$ resolution with a frame rate of 30 and 15 fps for day and night videos, respectively. For each subject, videos were recorded in a controlled setting in five conditions: (1) without glasses, (2) with glasses, (3) with sunglasses, (4) without glasses at night and (5) with glasses at night. Simulated behaviours include yawning, nodding, looking aside, talking, laughing, closing eyes and regular driving; video segments have been labelled as drowsy or non-drowsy. The dataset consists of training (18 persons), evaluation (4 persons) and testing (14 persons) sets. For this study, the training dataset was used for model calibration (a total of 8.5 h of video), the evaluation dataset for validation purposes (1.5 h of video), while the testing dataset was not used.

First, night videos were converted from 15 to 30 fps to match the frame rate of the other videos in the dataset. Videos were then resized from $480 \times 640$ to $240 \times 320$ to reduce pre-processing time during training and disc space. Using Python v3.6.1 and TensorFlow r1.4 [1], which provides optimisations for mobile inference, the video files were split into 100-frame sequences for training and 10-frame sequences for validation. This resulted in 9094 100-frame training records and 17,318 10-frame validation records, stored using TensorFlow's TFRecord format for increased read speed. Note that a small fraction of 10-frame sequences from the original videos is not used for training (i.e. 10-frame sequences spanning two records), which was a trade-off for faster read performance.

## 2.3 Pre-processing

Since the DDD dataset contains a limited amount of training data, several on-the-fly pre-processing steps were implemented to increase the variety of samples supplied to the neural network during training. These pre-processing steps were tailored to the issue of drowsiness detection and increase robustness of the model when applied in a real-world setting. First, ten consecutive frames were randomly selected from the 100-frame training record. This sequence was then flipped horizontally with a probability of 0.5, and the brightness was distorted by adding a single random value from $[-48/255, 48/255]$ to each pixel, before clipping to $[0, 1]$. These steps achieve model invariance with respect to (1) left-hand side versus right-hand side driving (given that the training videos are recorded from a fixed angle) and (2) varying brightness conditions inside a vehicle.

Figure 1 illustrates the remaining pre-processing steps. First, the sample was sliced using a randomly constructed inner box (dashed line), which was then rescaled to $224 \times 224 \times 10 \times 1$. The $x$ and $y$ dimensions of the inner box were randomly selected, subject to the following restrictions:

1. $x \leq 240$;

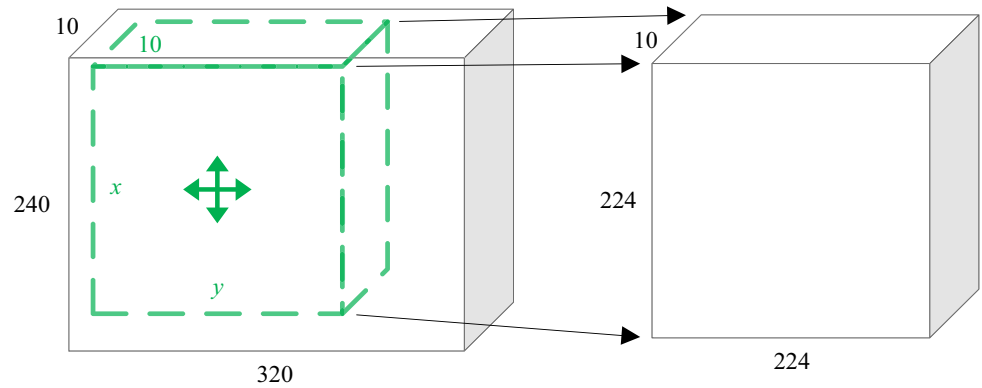

**Fig. 1** Pre-processing steps implementing translations, zooming and stretching

2. $\frac{xy}{240 \times 320} \geq 0.55$; i.e. the selected area should cover at least 55% of the original frame. This restricted zooming ensures that features important for drowsiness detection are unlikely to be removed by cropping; and

3. $y / x$ in [0.96, 1.04]; i.e. aspect ratio distortion corresponding to a maximum stretching of about $\pm 4\%$ in one dimension.

This approach achieves model invariance to translations (horizontal, vertical shifts), zooming, and face shapes. For example, minor stretching of the faces of the 18 persons in the training dataset covers a more varied range of face shapes, improving drowsiness detection on new drivers.

To illustrate the pre-processing steps described above, Fig. 2 presents various randomly distorted samples generated from the same input sequence of the DDD dataset, showing the onset of a nodding event. Row A includes distortions of the brightness level, zooming and a horizontal shift of the face position to the right, row B shows a darker sequence with a flip and vertical shift of the face position, and row C exemplifies stretching plus a flip.

## 2.4 Neural network architecture

Various architectures and parameter settings were explored in this research; the final model architecture that was selected will be described in this section. In addition, three existing network architectures were used as benchmarks to evaluate the differences in performance between 2D (frame-based) and 3D (video-based) models. The performance impact of adding requirements for mobile deployment was also explored; these two categories are referred to as 'Mobile deployment' (i.e. should be able to run real-time assessment on a mobile phone) and 'High compute' (i.e. without additional restrictions on computations). The selection of these benchmarks aimed to compare 2D versus 3D and high compute versus mobile deployment, ceteris paribus. Therefore, all models use the same $224 \times 224$ spatial resolution, while the video-based architectures are the 3D equivalent of the corresponding 2D model. The

Two-Stream Inflated 3D ConvNet (I3D) model [6] was used as a 3D, high compute benchmark model. For a fair comparison, we did not use the optical flow stream as motion extraction limits real-time assessment on mobile devices. I3D is based on InceptionV1 [44] with kernels inflated from 2D to 3D. Therefore, we state performance of the relatively older InceptionV1 architecture for the 2D, high compute category. This allows for a direct comparison of the impact of 2D and 3D convolutions on prediction accuracy. As benchmark 2D model for mobile deployment, we used the MobileNetV2 architecture [37]. MobileNetV2 was developed more recently than InceptionV1 and architectural improvements have not only resulted in lower computational requirements, but also higher accuracy. For example, the top-1 accuracies on ImageNet are 69.8% and 74.9% for InceptionV1 and MobileNetV2_1.4, respectively, when compared using similar setups [40]. The model architecture used in our research, based on MobileNetV2_1.4, aims to take advantage of performance improvements associated with (computationally expensive) 3D convolutions that integrate temporal information, while at the same time allowing for deployment on mobile devices.

Table 2 illustrates architectures for MobileNetV2_1.4 and our model. Using standard 3D convolution (conv3d) with filter size [3, 3, 3] in the first layer and depthwise separable 3D convolution (bottleneck3d) in the second layer, our model architecture results in an early fusion of spatial and temporal information. The bottleneck3d building block is an extension of the bottleneck residual block in [37], and a specification is given in Table 3. A kernel size of [3, 3, 5] is used in the depthwise operation of the bottleneck3d block to integrate all remaining temporal information. The two classification nodes represent the occurrence or absence of drowsiness.

## 2.5 Model calibration

All models were pre-trained on the ImageNet dataset, improving the capacity to process spatial information.

**Fig. 2** Random distortions of a 10-frame sequence; only frames 1, 2 and 10 are displayed. The top row shows the original sequence, while rows A, B and C are different randomly generated pre-processed samples based on the original sequence

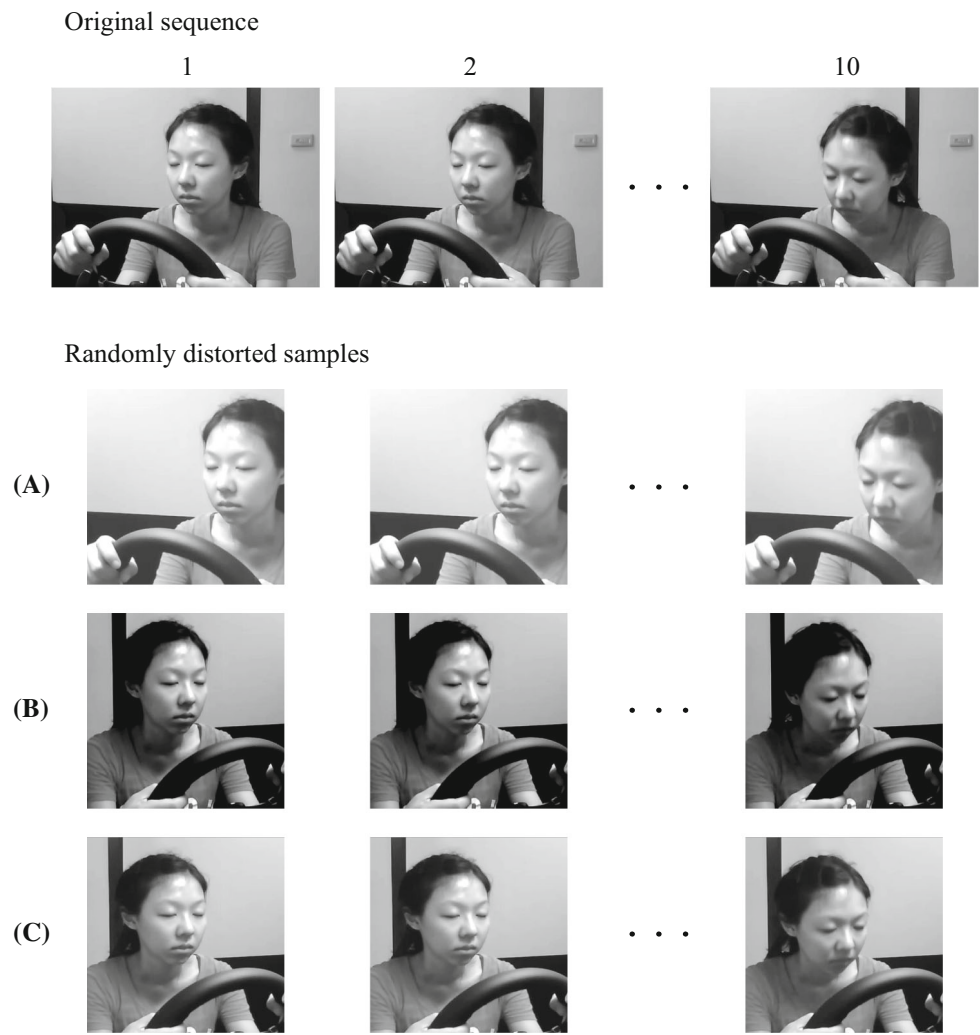

Original sequence

1                    2                    10

Randomly distorted samples

(A)

(B)

(C)

Video-based models may need additional pre-training for processing temporal information, as activity prediction can be more challenging than image recognition, since videos also include variations in motion and viewpoints [10]. Therefore, the data requirements for model calibration are likely to be more stringent. However, the DDD video dataset used in this study is small. Therefore, video-based models were pre-trained on the Kinetics-400 video database, containing 400 human action classes with over 400 clips per action. Although other video datasets could be used for pre-training as well, they either contain fewer classes (e.g. HMDB-51, UCF-101) or actions unrelated to drowsiness detection (e.g. Sports-1M [21]). In contrast, Kinetics' categories include 'yawning', 'laughing', 'shaking head' and 'baby waking up', which could capture features transferable to drowsiness detection. A version of the I3D model was used that had already been pre-trained on the Kinetics-400 video database. Following pre-training on ImageNet, our model was pre-trained on a small part of this Kinetics dataset for over 30 million iterations (approx. 3.5 days), using four NVIDIA P100 GPUs on a high-performance computing cluster at the University of Melbourne. Further performance improvements might be obtained by pre-training on the full Kinetics dataset. Weights were calibrated through supervised learning using an initial learning rate of 0.01, decaying to a minimum learning rate of 0.0001, momentum 0.9, weight decay of $4\text{E}-05$ and batch normalisation. The model was then fine-tuned on the DDD dataset for 10 million iterations (approx. 1 day), using a reduced initial learning rate (0.005 instead of 0.01) and reduced weight decay ($1\text{E}-07$ instead of $4\text{E}-05$).

## 2.6 Phone application

Faster inference speed and a smaller memory footprint were achieved by integrating the temporal dimension in the first few layers of the neural network. Further, operations not required for inference were stripped from the network and model weights were frozen [50]. The resulting model was deployed as an Android phone application using

**Table 2** MobileNetV2 [37] with width multiplier 1.4 (left) and our model architecture (right)

| Input dimension | Operator | $t$ | $c$ | $n$ | $s$ | Input dimension | Operator | $t$ | $c$ | $n$ | $s$ |
|---|---|---|---|---|---|---|---|---|---|---|---|
| *MobileNetV2_1.4* | | | | | | *Ours* | | | | | |
| | | | | | | $224^2 \times 10 \times 1$ | **conv3d** | – | **48** | 1 | **2** |
| $224^2 \times 3$ | conv2d | – | 48 | 1 | 2 | $112^2 \times 5 \times 48$ | **bottleneck3d** | 1 | **24** | 1 | **[1, 1, 5]** |
| $112^2 \times 48$ | bottleneck | 1 | 24 | 1 | 1 | $112^2 \times 1 \times 24$ | **squeeze** | – | – | 1 | – |
| $112^2 \times 24$ | bottleneck | 6 | 32 | 2 | 2 | $112^2 \times 24$ | bottleneck | 6 | 32 | 2 | 2 |
| $56^2 \times 32$ | bottleneck | 6 | 48 | 3 | 2 | $56^2 \times 32$ | bottleneck | 6 | 48 | 3 | 2 |
| $28^2 \times 48$ | bottleneck | 6 | 88 | 4 | 2 | $28^2 \times 48$ | bottleneck | 6 | 88 | 4 | 2 |
| $14^2 \times 88$ | bottleneck | 6 | 136 | 3 | 1 | $14^2 \times 88$ | bottleneck | 6 | 136 | 3 | 1 |
| $14^2 \times 136$ | bottleneck | 6 | 224 | 3 | 2 | $14^2 \times 136$ | bottleneck | 6 | 224 | 3 | 2 |
| $7^2 \times 224$ | bottleneck | 6 | 448 | 1 | 1 | $7^2 \times 224$ | bottleneck | 6 | 448 | 1 | 1 |
| $7^2 \times 448$ | conv2d $1 \times 1$ | – | 1792 | 1 | 1 | $7^2 \times 448$ | conv2d $1 \times 1$ | – | 1792 | 1 | 1 |
| $7^2 \times 1792$ | avgpool $7 \times 7$ | – | – | 1 | – | $7^2 \times 1792$ | avgpool $7 \times 7$ | – | – | 1 | – |
| $1 \times 1 \times 1280$ | conv2d $1 \times 1$ | – | 2 | – | | $1 \times 1 \times 1280$ | conv2d $1 \times 1$ | – | 2 | – | |

Parameters include the expansion factor in the bottleneck residual block ($t$) and the number of output channels ($c$). Layers are repeated $n$ times, using stride $s$ the first time and stride 1 afterwards. Differences with MobileNetV2 have been highlighted

**Table 3** 3D bottleneck residual block, transforming input of size height ($h$) × width ($w$) × number of frames ($f$) × input channels ($k$) using 3D depthwise separable convolution

| Input dimension | Operator | Output dimension |
|---|---|---|
| $h \times w \times f \times k$ | $1 \times 1 \times 1$ conv3d, ReLU6 | $h \times w \times f \times tk$ |
| $h \times w \times f \times tk$ | Depthwise with stride $[s_1, s_2, s_3]$, ReLU6 | $\frac{h}{s_1} \times \frac{w}{s_2} \times \frac{f}{s_3} \times tk$ |
| $\frac{h}{s_1} \times \frac{w}{s_2} \times \frac{f}{s_3} \times tk$ | Linear $1 \times 1 \times 1$ conv3d | $\frac{h}{s_1} \times \frac{w}{s_2} \times \frac{f}{s_3} \times c$ |

Android Studio, TensorFlow and Bazel (i.e. Google's open-source build tool). The application uses several parallel threads to (1) store an up-to-date stack of the latest ten frames of the front-facing camera, (2) perform inference using the trained model and (3) process inference results to warn a drowsy driver. Most of the computational capacity is spent on model inference; as soon as inference is completed, results processing starts in a separate thread and the latest ten frames are immediately used for a new inference run. The application plays a high-pitched warning sound based on the drowsiness probability estimate (i.e. > 50% probability). For demonstration purposes, a low-pitched sound is played at every inference otherwise. Thresholds for providing warning sounds could be customised; for example, increasing the detection threshold to 90% reduces the number of false positives, potentially improving user experience of the model.

## 3 Results

A performance comparison of the various architectures is presented in Table 4. Note that some studies report accuracy based on the DDD testing data, while training on the DDD training and evaluation datasets. As the testing data were not available to us, the out-of-sample performance reported here is for the DDD evaluation dataset (using the DDD training dataset for model calibration). All experiments use the same random distortions during pre-processing (see Sect. 2.3).

Similar to results on ImageNet, MobileNetV2_1.4 outperforms InceptionV1, due to advances in neural network architectures over time. The prediction accuracy of the I3D model is 5.8% higher than InceptionV1, which clearly highlights the benefits of using 3D over 2D convolutions for action recognition. This also applies to our model, which shows an increase in accuracy of 2.1% compared to MobileNetV2_1.4.

Detailed results, where performance is subdivided based on time of day and whether the driver is wearing (sun)-glasses, are presented in Table 5. In line with expectations, drowsiness is the easiest to predict when drivers are not wearing glasses (either during the day or at night). Further, Table 5 shows that the models based on 3D convolution operations perform better than the frame-based methods when a driver wears sunglasses (i.e. 55.9 and 59.3% for 2D vs. 74.7 and 76.8% for 3D). In this case, important information on eyelid closure is hidden from the image, leading to reduced performance for the frame-based methods. In contrast, the methods based on action recognition (i.e. I3D and ours) can maintain similar performance for drowsiness detection based on movement information embedded in the video.

**Table 4** Accuracy on the DDD evaluation dataset per deployment category and modelling approach

| | Frame (2D) | | Video (3D) | |
|---|---|---|---|---|
| | Model | Accuracy | Model | Accuracy |
| High compute | InceptionV1, Szegedy et al. [44] | 69.6% | I3D, Carreira and Zisserman [6] | 75.4% |
| Mobile deployment | MobileNetV2_1.4, Sandler et al. [37] | 71.8% | Ours | 73.9% |

**Table 5** Accuracy (%) of various methodologies and human labelling on the DDD evaluation dataset

| Scenario | High compute | | Mobile deployment | | Human accuracy |
|---|---|---|---|---|---|
| | InceptionV1 | I3D | MobileNetV2_1.4 | Ours | |
| No glasses | 76.0 | 78.9 | 74.3 | 75.4 | 82.0 |
| Glasses | 70.2 | 65.7 | 77.7 | 77.4 | 78.8 |
| Sunglasses | 55.9 | 74.7 | 59.3 | 76.8 | 80.9 |
| Night—no glasses | 73.0 | 79.7 | 73.8 | 76.1 | 82.5 |
| Night—glasses | 68.8 | 76.9 | 71.9 | 63.6 | 79.9 |
| All | 69.6 | 75.4 | 71.8 | 73.9 | 80.8 |

Park et al. [30] provide human performance statistics for the DDD evaluation dataset, where frames were manually labelled by doctoral students in computer vision. For sole drivers, a drowsiness detection system that has the same performance as a human observer could be compared to having someone continuously monitor their driving state even while driving alone (note that a detection system has consistent performance over longer periods). During daytime monitoring (i.e. no glasses, glasses, sunglasses), our method approaches human performance on drowsiness detection, exemplifying its potential application. Further, daytime results are favourable compared to the other tested models. The only area where our model does not perform very well is when a driver wears glasses at night. Interestingly, I3D has lower performance when a driver wears glasses during the day.

Due to changes to the neural network architecture for inference time optimisation, overall accuracy decreased by 1.5% compared to the I3D benchmark. However, this does allow for model deployment as a mobile phone application. Videos demonstrating this on a Samsung Galaxy S7 can be found as supplementary materials on the journal's website (Online Resources 1 and 2). Note that in these videos, the face of the first author was not part of the training dataset.

## 3.1 Sensitivity analysis

In the sensitivity analyses below, we present the impact of various assumptions on model performance. In each scenario described in Table 6, we vary only one parameter compared to the model presented in Sect. 2 and quantify its impact on accuracy. For each experiment related to the first four assumptions, the revised model was pre-trained on ImageNet and Kinetics, followed by fine-tuning on DDD. The sensitivity analyses for the final three assumptions used the same pre-trained model for all experiments. Some of these assumptions and model architectures also have implications on mobile inference time, which is presented in Table 7.

### 3.1.1 Sample length

Table 6 shows that accuracy can be further improved to 77.6% by increasing the length of input samples to 30 frames. Using 60 frames also led to better accuracy, although the early fusion architecture may be sub-optimal due to very large filter sizes. However, to respond to safety-critical events, timely detection is a key requirement of the model. The larger the input sample is, the longer the required recording time becomes before the inference process can start. Specifically, at 30 fps, video recording times are 0.17, 0.33, 1.0 and 2.0 s for 5, 10, 30 and 60 frames, respectively. This is not incorporated in the inference times stated in Table 7. In addition, the app did not run smoothly on our test device for input samples of 30 and 60 frames, presumably due to memory requirements for processing large sequences of frames. The neural network architectures in these experiments are similar because of the early fusion of spatial and temporal information; only the first few layers are different. Therefore, not much difference in inference time was observed between 5- and 10-frame sequences (see Table 7), although the 10-frame sequences yield substantially higher accuracy. Therefore,

**Table 6** Accuracy (%) on the DDD evaluation dataset, highlighting the **highest** accuracy and <u>selected</u> parameter value

| Assumption | Impact (reported as "accuracy (parameter value)") | | | | |
|---|---|---|---|---|---|
| Length of input sample (frames) | 65.9 (5) | 73.9 (<u>10</u>) | **77.6** (30) | 75.2 (60) | |
| Pre-processing steps: random flip, brightness (light), translate/zoom/stretch (tzs) | 64.0 (none) | 66.0 (flip) | 66.2 (light) | 68.7 (tzs) | **73.9** (<u>all</u>) |
| Fusion of spatial and temporal information | 73.9 (<u>early</u>) | **74.4** (late) | 71.0 (slow)[a] | | |
| Depth multiplier in MobileNetV2 | 68.9 (0.35) | 70.9 (0.75) | **73.9** (<u>1.4</u>) | | |
| Pre-training on ImageNet (IN), Kinetics (K), IN&K, or no pre-training (none) | 68.8 (none) | 69.6 (IN) | 73.6 (K) | **73.9** (<u>IN&K</u>) | |
| Fine-tuning: final layer only (final), all except early layers (most), or all layers | 50.5 (final) | 68.8 (most) | **73.9** (<u>all</u>) | | |
| Initial learning rate for fine-tuning | 65.5 (0.001) | **73.9** (<u>0.005</u>) | 71.6 (0.01) | | |
| Weight decay for fine-tuning | 70.9 (0) | **73.9** (<u>1E−07</u>) | 72.0 (4E−05) | | |

[a]The slow fusion architecture could not be pre-trained to the same extent as in other experiments due to high computational requirements

**Table 7** Inference time (s) on Samsung Galaxy S7, highlighting the **shortest** inference time and <u>selected</u> parameter value

| Assumption | Impact (reported as "inference time (parameter value)") | | | |
|---|---|---|---|---|
| Length of input sample (frames) | **1.0** (5) | 1.1 (<u>10</u>) | 1.9 (30) | — (60) |
| Fusion of spatial and temporal information | **1.1** (<u>early</u>) | 5.6 (late) | 31.7 (slow) | |
| Depth multiplier in MobileNetV2 | **0.5** (0.35) | 0.8 (0.75) | 1.1 (<u>1.4</u>) | |

using 10 frames provided a good balance between timely and high-quality predictions.

### 3.1.2 Pre-processing steps

The selected pre-processing steps all lead to an increase in prediction accuracy, individually. The largest increase in performance was observed for translations, zooming and stretching. Combining all pre-processing steps in the final model led to the highest overall accuracy.

### 3.1.3 Information fusion

The late fusion architecture reuses weights in Mobile-NetV2 for each frame and fuses the processed frames at the final layer, while the slow fusion architecture uses 3D depthwise separable convolutions throughout the network. Late fusion shows a slight increase in prediction accuracy compared to the early fusion approach. However, as shown in Table 7, the early fusion architecture resulted in much faster inference times, making it the preferred approach for mobile deployment.

### 3.1.4 Depth multiplier

Although reducing the depth multiplier in MobileNetV2 from 1.4 to 0.75 or 0.35 leads to much faster inference times, both show a substantial drop in prediction accuracy as well.

### 3.1.5 Pre-training

Pre-training improves prediction accuracy. In particular, the Kinetics video dataset provides good performance since our architecture was designed for video inputs. However, ImageNet pre-training still provides benefits due to enhanced processing of spatial information. ImageNet pre-training followed by Kinetics pre-training resulted in the highest accuracy and was adopted in our study.

### 3.1.6 Fine-tuning

Various transfer learning approaches were investigated, using the DDD dataset for fine-tuning. The training loss did not converge when fine-tuning weights in the final layer only. Freezing weights of the first few layers showed better results, preserving the low-level feature encodings obtained from Kinetics, which transfer reasonably well to the DDD dataset. However, fine-tuning all layers worked best. When setting the learning rate too high, performance peaked early during fine-tuning and declined with prolonged training. In contrast, setting the initial learning rate too low resulted in insufficient adaptation to the DDD dataset. Overall, fine-tuning weights in all layers using a reduced learning rate and weight decay (i.e. to preserve valuable information stored in the model weights during pre-training) provided the highest accuracy.

## 4 Discussion

This research developed a new method for real-time video monitoring with a 3D convolutional neural network, providing early warning signals to a drowsy driver. The model, using full and depthwise separable 3D convolutions in the first few layers, leads to a 2.1% improvement over MobileNetV2_1.4 on a drowsiness detection dataset. Key features include the ability of mobile phone deployment and the many pre-processing steps that make the model invariant to brightness conditions, video recording angle (cf. left vs. right-hand side driving), horizontal and vertical shifts (not requiring the driver to be exactly in the centre of the video), zoom level, and face shapes. Further, most existing drowsiness detection methods adopt a multi-step approach; i.e. separate feature detection from single frames, followed by a prediction step where features over multiple frames are linked to drowsiness. In principle, our method has several advantages over these methodologies by using a single model design:

- Our method implicitly decides which features are important for drowsiness detection, rather than the developer having to pre-specify a limited set of features such as eyelid closure and mouth position, potentially missing important distinctions such as outer brow raises, frowning, chin raises and nose wrinkles [49]. The presented methodology is capable of capturing these features, given a sufficient quality of the data labels.
- A second advantage is that spatial and temporal information is merged directly. Three-dimensional convolution filters incorporate the time dimension, enabling separation of blinking versus micro-sleep, talking versus yawning and the identification of important face movements.
- A third advantage (and future direction, see below) is that the model can easily be trained to detect other tasks given data availability, such as different levels of distraction. Similar associations with crash risk apply to distraction (e.g. [24, 48]), as fatigue and distraction both influence crash risk by withdrawing a driver's attention from the driving task [3, 54]. Further, distraction-related crashes resulted in more than 3400 fatalities and 391,000 injuries in 2015 [29]. Distraction encompasses a broad range of behaviours, including passenger interaction, eyes tracking away from the road, adjusting or monitoring in-vehicle systems and consuming food or beverages. Some existing drowsiness detection models use face tracking and clipping as a first step, removing important features such as hand movements, while other models may need to extend the feature detection step to add features important for distraction detection. Our method uses the complete video frame and is suitable to directly incorporate distraction-related actions by training on labelled distraction-related videos.

Overall, the results of the sensitivity analyses support the assumptions and choices made during this research, showing how to convert existing 2D architectures for video-processing, while balancing accuracy and run-time on mobile platforms.

### 4.1 Limitations

Deep learning has requirements on the quantity and quality of data used to train a model. The academic dataset used in this study has limitations in both areas, as the training data only consist of 18 persons and frames were not labelled one by one, but large segments were allocated the same label. Further, other approaches may currently be faster since inference of 10-frame sequences requires substantial computation power. However, this research has demonstrated the feasibility of real-time detection through the development of a phone application. This experiment also demonstrates the effectiveness of the new method on out-of-sample data (see the demo videos on the journal's website), regardless of the data limitations.

### 4.2 Future directions

For robust real-time assessment, a balanced sample of in-vehicle video footage recorded during naturalistic driving studies (e.g. [33, 48]) would provide a good data source for the calibration of the presented deep neural network. Naturalistic driving studies generally produce sufficient amounts of labelled data for supervised learning (i.e. for both drowsiness and distraction).

With respect to fusing spatial and temporal information, Karpathy et al. [21] found that late fusion led to higher accuracy than early fusion on the Sports-1M datasets (in line with our finding). Slow fusion resulted in the largest performance improvement over the baseline, frame-based method. At the moment of this research, no native support for depthwise separable 3D convolutions was available in TensorFlow, which prompted us to implement this functionality ourselves, resulting in slower performance. Consequently, the slow fusion architecture could not be pre-trained to the same extent and could not be used for real-time monitoring (31.7 s inference time on our mobile phone, see Sect. 3.1). However, a native implementation of depthwise separable 3D convolutions could make a slow fusion architecture feasible in the future.

New neural network architectures will also continue to become available for image or action recognition on mobile devices. For example, Chen et al. [8] designed a multi-fibre network architecture which reduces the computational costs of 3D networks. Rather than 3D depthwise separable convolutions, an ensemble of lightweight networks (i.e. fibres) is used in combination with multiplexer modules to reduce computational requirements. Further, deep learning frameworks have been extended with platforms specifically targeting inference on mobile devices (e.g. TensorFlow Lite), containing further optimisations to reduce mobile inference times. There is also a trend to equip new mobile phones with chips optimised for artificial intelligence (e.g. [31]). These technological and scientific advances present opportunities to further improve real-time, deep learning-based drowsiness detection models for practical deployment in applications such as presented here.

## 4.3 Implications

Although our research focussed on vehicle-based transport, the new method could be used for fatigue detection across a range of domains. Monitoring is especially useful for attention-critical tasks, where reduced task performance caused by fatigue can have serious consequences. For example, consider the tasks performed by air traffic controllers, airline pilots, nuclear power reactor operators and operators of heavy or dangerous machinery. In any of these contexts, a drowsiness detection system has potential to prevent accidents and save lives. Note that the model can be fine-tuned using task-specific data.

Finally, this study should be considered in the light of the impending introduction of autonomous vehicles. While fully autonomous vehicles would remove the need for drowsiness detection models (besides waking a sleeping driver when reaching the destination), such models can still be useful in the transition period where vehicles are not able to handle every encountered situation. The driver's state is valuable information to assess whether the driver is able to take back control on short notice. Until the roll-out of fully autonomous vehicles is completed, driver drowsiness will continue to increase crash risk and cause substantial road trauma each year. A drowsiness detection method that is widely available as a phone application has the potential to reduce drowsiness-related motor vehicle crashes and consequent road trauma well into the future.

**Acknowledgements** This work was supported by the Australian Research Council [Grant No. LP150100680]; QBE; and the Transport Accident Commission. Further, this research was undertaken using the LIEF HPC-GPGPU Facility hosted at the University of Melbourne [Grant No. LE170100200]. M.S. is supported by a National Health and Medical Research Council (Australia) Fellowship [Grant No. APP1136250]. J.T. is supported by an Australian Research Council Discovery Early Career Research Award [Grant No. DE180101411].

The authors would like to acknowledge Katherine Scully for her assistance with the literature review, and the anonymous reviewers for their valuable feedback, which helped improve the quality of the original manuscript.

## Compliance with ethical standards